\documentclass[letterpaper]{article}
\usepackage[preprint]{aaai2027}
\usepackage[hyphens]{url}
\usepackage{graphicx}
\usepackage{natbib}
\usepackage{caption}
\usepackage{algorithm}
\usepackage{algorithmic}
\usepackage{amsmath}
\usepackage{amsfonts}
\usepackage{amssymb}
\usepackage{booktabs}

\newtheorem{definition}{Definition}

\title{WorkflowOps: Learning Agent Collaboration Priors\\for Multi-Agent Workflow Orchestration}

\author{
    Qi Cheng\textsuperscript{\rm 1},
    Shengyu Chen\textsuperscript{\rm 2},
    Wei Cheng\textsuperscript{\rm 3},
    Zhengzhang Chen\textsuperscript{\rm 3},\\
    Xiaowei Jia\textsuperscript{\rm 1},
    Haoyu Wang\textsuperscript{\rm 3},
    Haifeng Chen\textsuperscript{\rm 3}
}
\affiliations{
    \textsuperscript{\rm 1}Rutgers University, New Brunswick, NJ\\
    \textsuperscript{\rm 2}University of Pittsburgh, Pittsburgh, PA\\
    \textsuperscript{\rm 3}NEC Labs America, Princeton, NJ\\
    q.cheng@rutgers.edu, shc160@pitt.edu, xj159@cs.rutgers.edu,\\
    \{weicheng, zchen, haoyu, haifeng\}@nec-labs.com
}

\begin{document}

\maketitle

\begin{abstract}
Multi-agent systems are increasingly deployed for complex knowledge work, yet their orchestration layers remain largely memoryless: each new task is decomposed, assigned, and executed from scratch with no benefit from prior successful executions. We present \textsc{WorkflowOps}, a multi-agent workflow orchestration framework that learns agent collaboration priors from historical workflows and expands its agent pool on demand to cover new capability requirements. Our approach introduces three coupled mechanisms. First, a \emph{transition probability matrix} captures pairwise agent collaboration frequencies from past workflows and applies them as soft guidance during DAG workflow construction through intra-layer ordering optimization, probability-thresholded edge suggestion, and transitive reduction for parallelism maximization. Second, a \emph{sufficiency-driven agent creation loop} detects capability gaps via semantic matching scores, generates specialized agents through an LLM, and simultaneously injects them into the collaboration matrix, so that newly created agents are immediately usable with predicted collaboration priors. Third, a \emph{layered semantic matching} strategy uses pre-trained sentence embeddings for fast, deterministic capability matching as a first pass, invoking LLM verification only for low-confidence cases, thereby reducing LLM routing calls by over 80\% compared to pure-LLM approaches. We formalize the workflow construction problem as constrained optimization over a dependency-valid DAG guided by Markov collaboration priors and show that even lightweight frequency-based statistics, without any training, provide effective orchestration guidance. Experiments on mixed code, math, and question-answering suites show that \textsc{WorkflowOps} improves end-to-end pass rates over recent workflow-construction baselines, with the largest gains on structured, decomposable tasks where past agent handoff patterns transfer.
\end{abstract}

\section{Introduction}
\label{sec:introduction}

\begin{figure*}[t]
  \centering
  \includegraphics[width=0.95\textwidth]{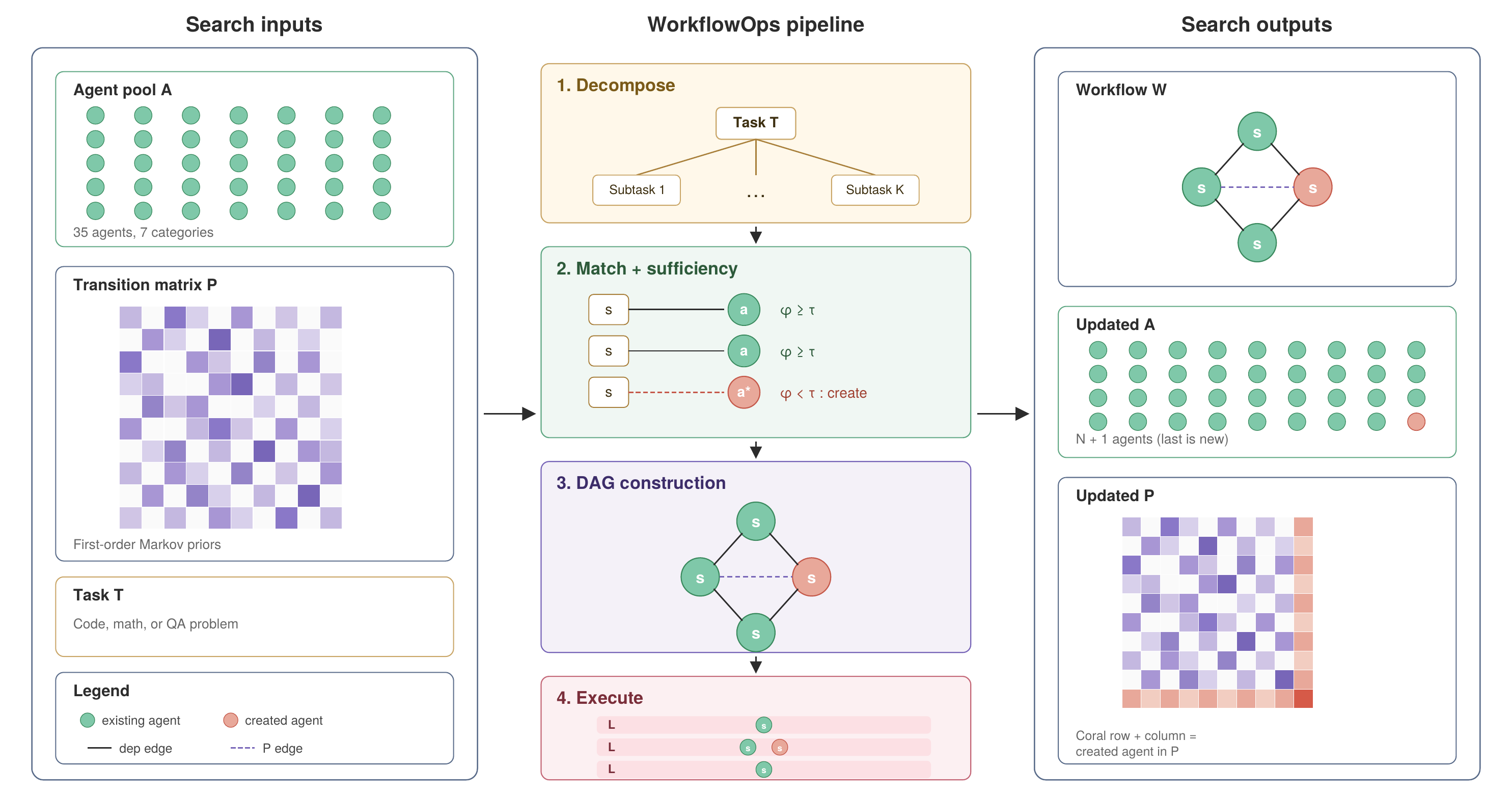}
  \caption{\textbf{Overview of \textsc{WorkflowOps}.} Given an agent pool $\mathcal{A}$, transition matrix $P$, and task $T$, the pipeline decomposes $T$, matches subtasks to agents (creating $a^{*}$ when $\phi<\tau$), constructs a DAG using $P$ as soft guidance, and executes it in parallel layers. Outputs: workflow $W$, expanded $\mathcal{A}'$, and updated $P'$. Coral marks the newly created agent.}
  \label{fig:workflowops_overview}
\end{figure*}

Large language model (LLM) based multi-agent systems have rapidly evolved from research prototypes into practical frameworks for complex knowledge-work tasks. Systems such as AutoGen~\citep{wu2024autogen}, CrewAI~\citep{crewai2024}, and LangGraph~\citep{langgraph2024} allow developers to compose specialized agents into collaborative workflows for tasks such as research synthesis, software engineering, compliance analysis, and business planning. As these systems are applied to increasingly complex and recurring tasks, the orchestration layer---how one task is decomposed, how subtasks are assigned to agents, how dependencies are structured, and how the resulting workflow is executed---has become a central bottleneck. Reconstructing a workflow from scratch for every query increases API cost, latency, and the risk of error accumulation, even when similar workflows have succeeded before.

The central limitation of many current orchestration pipelines is the lack of \emph{persistent orchestration state}: workflow-level collaboration priors, evolving capability coverage, and low-cost capability matching are rarely carried across tasks. Historical workflow traces expose recurring handoffs, capability gaps, and matching patterns that can be reused as soft guidance rather than hard constraints. Recent work improves orchestration adaptivity through topology selection, dynamic coordination, learned optimization, routing, and sub-agent creation~\citep{adaptorch2026,coral2026,flowsteer2026,masrouter2025,evoroute2026,aorchestra2026}. However, these threads are typically studied separately: some methods recompute decisions at task time, some require task-specific training, and others create agents without integrating them into a persistent collaboration model.

We introduce \textsc{WorkflowOps}, a history-conditioned orchestration framework that maintains persistent state through a collaboration-prior matrix, an evolving agent pool, and a semantic capability index. The system uses historical or bootstrapped successful workflow traces as soft guidance for approximate constrained optimization over dependency-valid DAGs, balancing historical alignment, assignment sufficiency, and structural parallelism. Concretely, \textsc{WorkflowOps} builds a row-stochastic agent transition matrix for layer ordering, soft-edge proposal, and redundancy pruning via transitive reduction~\citep{aho1972transitive}; expands the agent pool only when semantic sufficiency checks reveal capability gaps; and uses Sentence-BERT-style embeddings~\citep{reimers2019sbert} as the primary routing mechanism, with optional LLM verification for low-confidence cases.

We evaluate \textsc{WorkflowOps} on two heterogeneous suites: a Mixed code/math/QA benchmark and a harder Mixed\_Hard suite of competition-grade and frontier-difficulty problems. On the Mixed test split, \textsc{WorkflowOps} in DAG mode achieves a 0.890 overall pass rate (Linear: 0.870), improving over MetaAgent~\citep{metaagent}, the strongest query-level baseline, by 13.9 absolute points, and over ADAS~\citep{adas}, the strongest task-level baseline, by 19.6 points. On the harder Mixed\_Hard suite, \textsc{WorkflowOps} reaches 0.770, exceeding every fully-completed baseline by at least 15 points. Ablations show that removing the transition matrix, disabling agent-pool evolution, and replacing semantic embeddings with TF-IDF reduce accuracy by 4.5, 1.7, and 8.5 points, respectively, validating each of the three coupled mechanisms. The gains are strongest on code and math; DAG execution provides modest additional accuracy gains via matrix-suggested edges over the Linear mode.

Our contributions are as follows:

\begin{enumerate}
    \item We formulate history-conditioned multi-agent workflow construction as approximate constrained optimization over dependency-valid DAGs, guided by persistent Markov collaboration priors extracted from successful workflow traces.

    \item We instantiate this formulation with a transition-prior workflow-construction algorithm that uses empirical agent-to-agent transition probabilities for layer ordering, soft-edge proposal, and redundancy pruning, while preserving hard task dependencies.

    \item We introduce a sufficiency-driven agent-pool expansion loop that detects capability gaps through semantic matching, creates specialized agents only when needed, and immediately integrates new agents into the collaboration-prior matrix.

    \item We provide an end-to-end empirical study and ablation analysis on Mixed and Mixed\_Hard benchmarks, showing a 13.9-point overall gain over the strongest query-level baseline on Mixed and a 15+ point lead over fully-completed baselines on Mixed\_Hard, while isolating the effects of transition priors, agent evolution, and semantic matching.
\end{enumerate}

Together, these results position \textsc{WorkflowOps} as a lightweight and interpretable framework for training-free, history-conditioned multi-agent orchestration. Rather than relying on expensive policy learning or reconstructing workflows from scratch, \textsc{WorkflowOps} demonstrates that simple frequency-based collaboration statistics, when coupled with persistent agent-pool evolution and semantic capability matching, can serve as effective operational priors for future workflow construction.

\section{Related Work}
\label{sec:related_work}

Recent multi-agent orchestration methods improve adaptivity through topology selection, dynamic coordination, learned workflow optimization, routing, or agent creation. AdaptOrch~\citep{adaptorch2026} selects among predefined orchestration patterns, CORAL~\citep{coral2026} studies information-flow-based dynamic coordination, and FlowSteer~\citep{flowsteer2026} learns workflow-editing policies through reinforcement learning. Other work optimizes workflow structure or interaction policies directly, including HeGFlow~\citep{hegflow2025}, AgentPO~\citep{agentpo2026}, and agentic architecture search methods such as ADAS~\citep{adas} and MaAS~\citep{maas}. These approaches show that orchestration structure matters, but many either recompute decisions at task time or require training/search for each target setting.

Our baselines instantiate this workflow-search line: AFlow~\citep{aflow} searches workflow code with MCTS, ADAS~\citep{adas} performs automated agentic-system search, MaAS~\citep{maas} searches over architecture supernets, and MetaAgent~\citep{metaagent} constructs a finite-state-machine workflow per query. A further line reuses workflow \emph{experience}: Agent Workflow Memory~\citep{awm2024} induces and stores reusable workflow skills, EvoFlow~\citep{evoflow2025} evolves a population of diverse workflows, and TopoPrior~\citep{topoprior2026} learns topology priors that transfer across domains. \textsc{WorkflowOps} is complementary to both lines: rather than storing executable workflows or learning topology generators, it accumulates interpretable agent-to-agent transition statistics and applies them as training-free soft guidance under hard task dependencies.

A complementary line of work studies agent routing and self-evolving agent systems. MasRouter~\citep{masrouter2025} and EvoRoute~\citep{evoroute2026} learn routing policies for multi-agent systems, while AOrchestra~\citep{aorchestra2026}, SEAA~\citep{seaa2026}, SEVerA~\citep{severa2026}, and AgentOrchestra~\citep{agentorchestra2025} explore dynamic sub-agent creation, self-extension, verification, or hierarchical orchestration. \textsc{WorkflowOps} differs by integrating these threads into a single training-free, history-conditioned orchestration layer: empirical transition priors guide DAG construction, semantic matching assigns subtasks to agents, and demand-driven agent creation expands the pool while immediately updating the persistent collaboration prior. Extended related work is provided in Appendix~\ref{app:related-work}.

\section{Problem Formulation}
\label{sec:problem}

\paragraph{General problem.} Given a task $T$ and an agent pool $\mathcal{A}$, workflow orchestration must produce a dependency-valid workflow DAG $W$ and an agent assignment $\pi$ that maximize downstream task success. This statement is independent of any particular mechanism. \textsc{WorkflowOps} additionally assumes access to a history of successful workflows, from which it derives a transition matrix $\mathbf{P}$; $\mathbf{P}$ is an input to \emph{our approach}, not part of the problem itself. We first define the core objects.

\begin{definition}[Agent Pool]
An agent pool $\mathcal{A}=\{a_1,\ldots,a_N\}$ consists of $N$ specialized agents. Each agent $a_i$ is characterized by a natural language capability description $d_i$ and dense embedding $\mathbf{e}_i=f_{\mathrm{enc}}(d_i)\in\mathbb{R}^h$, where $f_{\mathrm{enc}}$ is a pre-trained sentence encoder.
\end{definition}

\begin{definition}[Task Decomposition]
Given a natural language task $T$, a decomposition function $\mathcal{D}_{\mathrm{LLM}}$ produces subtasks $\mathcal{S}=\{s_1,\ldots,s_K\}$ and a dependency DAG $G_{\mathrm{dep}}=(\mathcal{S},E_{\mathrm{dep}})$, where each directed edge $(s_i,s_j)\in E_{\mathrm{dep}}$ indicates that $s_i$ must complete before $s_j$ begins. Each subtask $s_j$ carries a required capability description $c_j$.
\end{definition}

\begin{definition}[Capability Matching]
\label{def:matching}
For subtask $s_j$ with requirement $c_j$, the matching score against agent $a_i$ is
\begin{equation}
\phi(s_j,a_i)=\cos\!\big(f_{\mathrm{enc}}(c_j),\mathbf{e}_i\big).
\end{equation}
The assignment $\pi:\mathcal{S}\to\mathcal{A}$ maps each subtask to its best-matching agent: $\pi(s_j)=\arg\max_{a_i\in\mathcal{A}}\phi(s_j,a_i)$.
\end{definition}

\begin{definition}[Sufficiency]
\label{def:sufficiency}
Assignment $\pi$ is $\tau$-sufficient if and only if $\phi(s_j,\pi(s_j))\geq \tau$ for all $s_j\in\mathcal{S}$, where $\tau\in[0,1]$ is the sufficiency threshold. The gap set is $\mathcal{G}=\{s_j\mid \phi(s_j,\pi(s_j))<\tau\}$.
\end{definition}

\begin{definition}[Transition Matrix]
\label{def:transition_matrix}
Given a history of successful workflows $\mathcal{H}=\{W_1,\ldots,W_M\}$, each containing an ordered sequence of agent invocations, the transition matrix $\mathbf{P}\in\mathbb{R}^{N\times N}$ is
\begin{equation}
P_{ij}=\frac{\mathrm{count}(a_i\to a_j\text{ in }\mathcal{H})}{\sum_{k=1}^{N}\mathrm{count}(a_i\to a_k\text{ in }\mathcal{H})}.
\end{equation}
Here $\mathrm{count}(a_i\to a_j\text{ in }\mathcal{H})$ denotes the number of times agent $a_j$ immediately follows $a_i$ across workflows in $\mathcal{H}$. $\mathbf{P}$ is row-stochastic and encodes empirical first-order Markov collaboration priors.
\end{definition}

\begin{definition}[Workflow]
A workflow $W=(V,E,\pi)$ consists of nodes $V=\mathcal{S}$, edges $E=E_{\mathrm{dep}}\cup E_{\mathrm{mat}}$ containing hard dependency edges and soft matrix-suggested edges, and agent assignment $\pi$.
\end{definition}

\paragraph{WorkflowOps design objective.} Because downstream task success is not directly measurable at construction time, our approach optimizes a history-conditioned surrogate balancing three terms: alignment with historical collaboration priors, semantic sufficiency of subtask--agent assignments, and structural parallelism:
\begin{equation}
\label{eq:objective}
Q(W, \pi, \mathcal{A}') =
\alpha \cdot \mathrm{Align}(W, \mathbf{P})
+ \beta \cdot \mathrm{Suff}(\pi)
+ \gamma \cdot \mathrm{Par}(W),
\end{equation}
subject to preserving hard task dependencies, maintaining acyclicity, and allowing only monotonic expansion of the agent pool $\mathcal{A}' \supseteq \mathcal{A}$, where
\begin{align}
\mathrm{Align}(W,\mathbf{P}) &= \frac{1}{|E|}\sum_{(s_i,s_j)\in E} P_{\pi(s_i),\pi(s_j)}, \label{eq:align}\\
\mathrm{Suff}(\pi) &= \frac{1}{|\mathcal{S}|}\sum_{j=1}^{K}\phi(s_j,\pi(s_j)), \label{eq:suff}\\
\mathrm{Par}(W) &= \frac{|\mathcal{S}|}{\mathrm{depth}(W)}, \label{eq:par}
\end{align}
with $\mathrm{depth}(W)$ the length of the longest path in the DAG. Align measures agreement with historical collaboration patterns, Suff measures capability coverage, and Par measures available parallelism. We use this objective as a design principle rather than solving it globally; \textsc{WorkflowOps} approximates it through staged decomposition, semantic matching, agent-pool expansion, transition-guided DAG construction, and layered execution.

\section{Method}
\label{sec:method}

\textsc{WorkflowOps} operates as a pipeline of four stages: (1) task decomposition with layered semantic matching, (2) sufficiency-driven agent pool expansion, (3) transition-prior-guided workflow construction, and (4) layered parallel execution. We describe each stage below, with detailed algorithms where useful. The general flow of WorkflowOps can be found in Figure~\ref{fig:workflowops_overview}.

\subsection{Task decomposition and layered semantic matching}
\label{sec:matching}

\paragraph{Task decomposition.} Given a natural language task $T$, an LLM decomposes it into $K$ subtasks $\mathcal{S} = \{s_1, \ldots, s_K\}$ ($3 \leq K \leq 7$), each with a required capability description $c_j$ and dependency links forming a DAG $G_{\mathrm{dep}}$. The decomposition prompt enforces valid DAG structure; a post-processing step detects and breaks cycles via DFS if they arise from LLM output errors. All dependency links are thus proposed by the LLM under a schema-constrained prompt: structural validity is enforced programmatically, while the semantic correctness of the decomposition is delegated to the LLM.

\paragraph{Layered semantic matching.} We adopt a two-layer matching strategy that balances quality and efficiency:

\textbf{Layer 1: Sentence embedding matching.} We use a pre-trained sentence encoder $f_{\mathrm{enc}}$ (Sentence-BERT~\citep{reimers2019sbert}, specifically the \texttt{all-MiniLM-L6-v2} variant) to encode both agent capability descriptions and subtask requirements into 384-dimensional dense vectors. Matching scores are computed as:
\begin{equation}
\phi(s_j, a_i) = \cos\!\big(f_{\mathrm{enc}}(c_j),\; f_{\mathrm{enc}}(d_i)\big).
\end{equation}
Here $a_i$ denotes an agent, $d_i$ its capability description, and $c_j$ the subtask requirement (cf.\ Definition~\ref{def:matching}). Agent embeddings are pre-computed and cached; subtask embeddings are computed at decomposition time. This layer is deterministic, has zero LLM cost, and runs in sub-millisecond time per subtask-agent pair through batched matrix multiplication.

\textbf{Layer 2: LLM on-demand verification.} For subtasks where the best matching score falls below the sufficiency threshold $\tau$, an optional LLM verification stage can judge whether an existing agent is truly adequate or whether a capability gap exists. This stage is invoked only for low-confidence cases where embedding similarity is ambiguous.

\paragraph{Complexity comparison.} Pure LLM routing evaluates $K \times N$ subtask-agent pairs with LLM calls. Our approach requires $K \times N$ vector operations (Layer 1, negligible cost) plus at most $K' \times 1$ LLM calls (Layer 2, where $K' \ll K$). For a typical setting with $K{=}5$ subtasks and $N{=}35$ agents, this reduces from 175 LLM calls to ${\sim}5$ vector-similarity computations plus ${\sim}1$ LLM verification call.

\subsection{Transition-prior-guided workflow construction}
\label{sec:transition_matrix}

\paragraph{Matrix construction.} The transition matrix $\mathbf{P}$ (Definition~\ref{def:transition_matrix}) is built offline from a corpus of historical (or synthetic) workflows by counting pairwise agent transitions and row-normalizing. In our deployment, we bootstrap from ${\sim}65$ synthetic workflows spanning eight task domains, yielding a $35 \times 35$ matrix with approximately 20\% non-zero entries. Constructing $\mathbf{P}$ is a single counting pass over historical handoffs, lookups are $O(1)$, and memory grows as $O(N^2)$; at $N{=}35$ the matrix occupies a few kilobytes, so its construction and query cost are negligible relative to a single LLM call.

\paragraph{Triple-role mechanism.} The transition matrix serves three distinct roles during workflow construction:

\textbf{(a) Intra-layer ordering optimization.} Within each topological layer (subtasks that share the same depth and can be freely reordered), agents are sorted by their transition probability from the preceding layer's agents. If the previous layer's agent was $a_p$, and the current layer contains subtasks assigned to agents $a_x$ and $a_y$, the system prefers the ordering where $P_{p,x} > P_{p,y}$, placing $a_x$ before $a_y$. This optimizes information flow sequence based on historically successful patterns.

\textbf{(b) Probability-thresholded edge suggestion.} For each pair of subtasks $(s_i, s_j)$ with no explicit dependency edge, if $P_{\pi(s_i), \pi(s_j)} \geq \theta$ (the edge threshold) and adding the edge preserves topological ordering (i.e., $\mathrm{depth}(s_i) < \mathrm{depth}(s_j)$), a soft edge is added. These matrix-suggested edges encode the system's belief, grounded in historical data, that agent $\pi(s_i)$'s output is valuable as input to agent $\pi(s_j)$.

\textbf{(c) Transitive reduction.} After both hard dependency edges and soft matrix-suggested edges are incorporated, the system performs a transitive reduction~\citep{aho1972transitive}: if edges $(u, v)$ and $(v, w)$ exist, and a direct edge $(u, w)$ also exists, the direct edge $(u, w)$ is removed because the dependency is already transitively satisfied. This maximizes parallelism by removing redundant sequential constraints, allowing more subtasks to execute concurrently within each topological layer.

Algorithm~\ref{alg:dag_construction} formalizes this procedure.

\begin{algorithm}[t]
\small
\caption{Transition-guided DAG workflow construction}
\label{alg:dag_construction}
\begin{algorithmic}[1]
\REQUIRE Subtasks $\mathcal{S}$, dependency graph $G_{\mathrm{dep}} = (\mathcal{S}, E_{\mathrm{dep}})$, assignment $\pi$, transition matrix $\mathbf{P}$, edge threshold $\theta$
\ENSURE Workflow $W = (V, E, \pi)$
\STATE $V \leftarrow \mathcal{S}$; \quad $E \leftarrow E_{\mathrm{dep}}$
\STATE Compute topological depths: $\mathrm{depth}(s) \leftarrow$ longest path from any source to $s$ in $G_{\mathrm{dep}}$
\FOR{each pair $(s_i, s_j) \in \mathcal{S} \times \mathcal{S}$ with $s_i \neq s_j$ and $(s_i, s_j) \notin E$}
    \IF{$P_{\pi(s_i), \pi(s_j)} \geq \theta$ \AND $\mathrm{depth}(s_i) < \mathrm{depth}(s_j)$}
        \STATE $E \leftarrow E \cup \{(s_i, s_j)\}$ \hfill $\triangleright$ Add matrix-suggested soft edge
    \ENDIF
\ENDFOR
\STATE $E \leftarrow \textsc{TransitiveReduction}(V, E)$ \hfill $\triangleright$ Remove redundant edges
\STATE Compute parallel groups via layered topological sort of $(V, E)$
\STATE Sort agents within each layer by $P_{\mathrm{prev}, \pi(s)}$ descending \hfill $\triangleright$ Intra-layer ordering
\RETURN $W = (V, E, \pi)$
\end{algorithmic}
\end{algorithm}

\paragraph{Dynamic matrix update.} When new agents are created (\S\ref{sec:self_evolving}), they must be integrated into $\mathbf{P}$. The system uses the LLM to predict likely predecessor and successor agents for the new agent $a_{\mathrm{new}}$; the prediction is conditioned on the new agent's capability card and the capability descriptions of the existing pool (cf.\ Algorithm~\ref{alg:sufficiency}), and the predicted transition probabilities are injected into the corresponding rows and columns of $\mathbf{P}$. All affected rows are re-normalized to maintain row-stochasticity. This ensures that the matrix grows alongside the agent pool, so workflow construction can immediately exploit the new agent's predicted collaboration patterns.

\paragraph{Why soft guidance.} The matrix is used as soft guidance rather than hard constraints for three reasons: (i) with ${\sim}20\%$ matrix density, hard constraints would block valid but historically unseen workflows; (ii) dependency structure is always the non-negotiable foundation---the matrix cannot override task logic; (iii) novel task patterns should still be expressible even if they deviate from historical priors.

\subsection{Sufficiency-driven agent-pool expansion}
\label{sec:self_evolving}

When the current agent pool cannot $\tau$-sufficiently cover a task's requirements, \textsc{WorkflowOps} enters a closed-loop agent creation process. This is a defining capability: the system does not stop at weak assignments but actively expands its capability boundaries.

Algorithm~\ref{alg:sufficiency} formalizes the sufficiency loop.

\begin{algorithm}[t]
\small
\caption{Sufficiency-driven agent pool expansion}
\label{alg:sufficiency}
\begin{algorithmic}[1]
\REQUIRE Subtasks $\mathcal{S}$, agent pool $\mathcal{A}$, threshold $\tau$, max rounds $R$, matrix $\mathbf{P}$
\ENSURE Expanded pool $\mathcal{A}'$, updated matrix $\mathbf{P}'$, sufficient assignment $\pi'$
\STATE $\mathcal{A}' \leftarrow \mathcal{A}$; \quad $\mathbf{P}' \leftarrow \mathbf{P}$
\FOR{$r = 1, \ldots, R$}
    \STATE Build matching index from $\mathcal{A}'$ using $f_{\mathrm{enc}}$
    \STATE Compute $\pi' \leftarrow \arg\max_{a \in \mathcal{A}'} \phi(s_j, a)$ for all $s_j \in \mathcal{S}$
    \STATE $\mathcal{G} \leftarrow \{s_j \mid \phi(s_j, \pi'(s_j)) < \tau\}$ \hfill $\triangleright$ Identify capability gaps
    \IF{$\mathcal{G} = \emptyset$}
        \STATE \textbf{break} \hfill $\triangleright$ All subtasks sufficiently covered
    \ENDIF
    \FOR{each $s_j \in \mathcal{G}$}
        \STATE \textbf{if} optional adequacy check indicates existing agents are insufficient \textbf{then}
        \STATE \quad $a_{\mathrm{new}} \leftarrow \textsc{LLMGenerate}(s_j.\mathrm{capability})$ \hfill $\triangleright$ Generate agent definition
        \STATE \quad Persist $a_{\mathrm{new}}$ to agent pool on disk
        \STATE \quad $\mathcal{A}' \leftarrow \mathcal{A}' \cup \{a_{\mathrm{new}}\}$
        \STATE \quad $(\mathrm{pred}, \mathrm{succ}) \leftarrow \textsc{LLMPredict}(a_{\mathrm{new}}, \mathcal{A}')$ \hfill $\triangleright$ Predict collaborators
        \STATE \quad $\mathbf{P}' \leftarrow \textsc{InjectAndNormalize}(\mathbf{P}', a_{\mathrm{new}}, \mathrm{pred}, \mathrm{succ})$
    \ENDFOR
\ENDFOR
\RETURN $\mathcal{A}', \mathbf{P}', \pi'$
\end{algorithmic}
\end{algorithm}

\paragraph{Key design decisions.}

\textbf{Demand-driven creation.} Unlike AOrchestra~\citep{aorchestra2026}, which creates agents for every task regardless of pool coverage, \textsc{WorkflowOps} creates agents only when a measurable capability gap is detected ($\phi < \tau$). This prevents unnecessary pool inflation.

\textbf{Integrated collaboration knowledge.} Each new agent is not only added to the pool but also injected into the transition matrix $\mathbf{P}$ with LLM-predicted predecessor and successor relationships. This means the next workflow that uses the new agent benefits from collaboration priors immediately, rather than requiring multiple executions to accumulate transition statistics organically.

\textbf{Safety mechanisms.} The loop is bounded by three mechanisms: (i) a maximum round count $R$ (default 3), (ii) a per-subtask deduplication guard, and (iii) a maximum agent creation count per run.

\textbf{Persistence and lifecycle.} Created agents are persisted to disk and become part of the available pool for future runs. The system supports three environment modes---exploratory (free evolution), controlled (periodic curation), and frozen (version-locked for production)---enabling a governance path from experimentation to enterprise deployment.

\subsection{Layered Parallel Execution}
\label{sec:execution}

After workflow construction, \textsc{WorkflowOps} executes the DAG by topological layers. All nodes in the same layer are dependency-independent and can therefore run concurrently; downstream nodes receive either a single predecessor output or a merged message containing all upstream results. This execution mode preserves hard dependencies while exploiting the parallelism exposed by transition-guided DAG construction.

We compare this DAG mode with a Linear mode that uses the same decomposition and agent assignment but executes subtasks sequentially. Both modes use the same message abstraction and logging interface, enabling a controlled comparison between sequential execution and transition-guided DAG execution. Detailed message schemas, fan-in merging, error propagation, audit logging, and the end-to-end pipeline algorithm (Algorithm~\ref{alg:pipeline}) are provided in Appendix~\ref{app:execution-details}.

\section{Experiments}
\label{sec:experiments}

\subsection{Experimental Setup}
\label{sec:exp_setup}

\paragraph{Task suites.}
We use two heterogeneous benchmark suites with the same inference pipeline and grading harness. \textbf{Mixed} combines code generation (HumanEval~\citep{humaneval}, MBPP~\citep{mbpp}), mathematical reasoning (GSM8K~\citep{gsm8k}, MATH~\citep{math}), and question answering (DROP~\citep{drop}, HotpotQA~\citep{hotpotqa}); it contains 120 validation and 300 test problems, sampled equally across sources. \textbf{Mixed\_Hard} contains harder competition-grade and frontier-difficulty problems from AIME~2025~\citep{amine}, LiveCodeBench~\citep{livecodebench}, CodeContests~\citep{codecontest}, and GPQA Diamond~\citep{GPQA}; it contains 180 validation and 694 test problems. The orchestrator does not observe the benchmark identity, so it must handle both broad category coverage and harder problem distributions without source-specific routing.

\paragraph{Baselines.}
We compare against four recent multi-agent orchestration systems: AFlow~\citep{aflow}, an MCTS-based workflow generator; ADAS~\citep{adas}, an automated agentic-system search method; MaAS~\citep{maas}, architecture search over agentic supernets; and MetaAgent~\citep{metaagent}, a finite-state-machine multi-agent constructor. MetaAgent constructs a fresh workflow per query, while AFlow, ADAS, and MaAS search for one workflow per benchmark and reuse it across queries.

\paragraph{Backbone and configuration.}
All systems use \texttt{gpt-5} as the underlying language model. \textsc{WorkflowOps} uses sufficiency threshold $\tau{=}0.45$, edge threshold $\theta{=}0.3$, the \texttt{all-MiniLM-L6-v2} sentence-transformer matcher~\citep{reimers2019sbert}, \texttt{MAX\_AGENTS\_PER\_RUN}$=4$, \texttt{MAX\_CREATION\_ROUNDS}$=3$, and \texttt{MAX\_DAG\_BRANCHING}$=3$. Decomposition produces $3{-}7$ subtasks per query, workflows execute with 16 worker processes per configuration, and each query has a 900\,s wall-clock timeout.

\paragraph{Metrics.}
The primary metric is pass rate: the fraction of test rows where the extracted final answer matches ground truth, with code tasks graded by sandboxed execution against canonical tests. We also log matrix alignment, sufficiency, dynamic-agent creation, gap closure, LLM calls, token counts, latency, and DAG-vs.-Linear wall time; detailed operational statistics are reported in Appendix~\ref{app:efficiency}.

\subsection{End-to-end Workflow Quality}
\label{sec:exp_main}

Table~\ref{tab:main_results_compact} reports the headline pass rates on Mixed and Mixed\_Hard. Full category-level breakdowns are provided in Appendix~\ref{app:full-results}. Baseline numbers are taken from the respective systems' evaluation runs on the same splits with the same \texttt{gpt-5} backbone.

\begin{table}[t]
\centering
\small
\caption{End-to-end pass rate on Mixed and Mixed\_Hard. Full category breakdowns are in Appendix~\ref{app:full-results}.}
\label{tab:main_results_compact}
\begin{tabular}{@{}lcc@{}}
\toprule
System & Mixed & Mixed\_Hard \\
\midrule
AFlow~\citep{aflow} & 0.676 & 0.218 \\
ADAS~\citep{adas} & 0.694 & 0.729$^{\dagger}$ \\
MaAS~\citep{maas} & 0.400 & DNF \\
MetaAgent~\citep{metaagent} & 0.751 & 0.618 \\
\midrule
\textsc{WorkflowOps} Linear & 0.870 & 0.763 \\
\textsc{WorkflowOps} DAG & \textbf{0.890} & \textbf{0.770} \\
\bottomrule
\end{tabular}
\vspace{2pt}
\begin{flushleft}
\footnotesize $^{\dagger}$ADAS on Mixed\_Hard is the strongest partial-strategy result among completed candidate strategies. DNF = did not finish (MaAS deadlocked on Mixed\_Hard).
\end{flushleft}
\end{table}

\paragraph{Mixed.}
On Mixed, \textsc{WorkflowOps} (DAG) reaches $0.890$, improving over the strongest query-level baseline, MetaAgent ($0.751$), by $13.9$ absolute points and over the strongest task-level baseline, ADAS ($0.694$), by $19.6$ points. The gain is largest on CODE: \textsc{WorkflowOps} obtains $0.940$ in Linear mode and $0.960$ in DAG mode, more than $40$ points above the nearest baseline. On MATH, \textsc{WorkflowOps} ties MetaAgent at $0.940$. QA remains the weakest category: MetaAgent leads by $4$--$8$ points, suggesting that the current agent pool and transition priors are better aligned with structured code/math workflows than with open-ended reading-comprehension style reasoning. DAG mode improves over Linear by $2$ points overall, with the gain concentrated on CODE and QA.

\paragraph{Mixed\_Hard.}
On the harder benchmark suite, \textsc{WorkflowOps} (DAG) reaches $0.770$, outperforming every fully completed baseline by at least $15$ points and exceeding the strongest partial ADAS result by $4$ points. The breakdown shows strong performance across competition-grade problems: $0.850$ on MATH, $0.780$ on CODE, and $0.700$ on QA. DAG and Linear modes are within $1$ point of each other on Mixed\_Hard, indicating that the harder suite benefits more from robust assignment and agent coverage than from additional DAG structure.

\paragraph{Significance.}
A paired bootstrap test on Mixed finds the gap between \textsc{WorkflowOps} (DAG) and MetaAgent significant ($p{<}0.001$). The effect is driven primarily by CODE; MATH is statistically tied and QA marginally favors MetaAgent. Full confidence intervals, category results, and effect sizes are in Appendix~\ref{app:full-results}.

\subsection{Ablation Studies}
\label{sec:exp_ablation}

We ablate the mechanisms introduced in Section~\ref{sec:method}: transition-matrix guidance, sufficiency-driven agent-pool expansion, semantic embedding matching, pure-LLM matching, and the edge threshold $\theta$. Each cell uses the same agent pool, matrix initialization, decomposition prompt, and execution engine.

\begin{table*}[t]
\centering
\small
\caption{Ablation results on the Mixed test split (300 problems per cell). ``Linear'' and ``DAG'' columns give overall pass rate; brackets are 95\% bootstrap CIs ($n_\text{boot}{=}2000$). CODE/MATH/QA are unweighted category means in Linear mode. Calls and tokens are mean per query in Linear mode.}
\label{tab:ablation}
\begin{tabular}{l c c c c c | c c}
\toprule
Variant & Linear & DAG & CODE & MATH & QA & calls & tokens \\
\midrule
\textsc{Reference}    & \textbf{0.870} \scriptsize{[0.830, 0.907]} & \textbf{0.890} \scriptsize{[0.853, 0.923]} & \textbf{0.940} & 0.940 & \textbf{0.730} & 14.8 & 43.6k \\
\midrule
\textsc{No-Matrix}    & 0.825 \scriptsize{[0.780, 0.867]} & 0.840 \scriptsize{[0.797, 0.880]} & 0.920 & 0.910 & 0.645 & 13.9 & 41.2k \\
\textsc{No-Evolve}    & 0.853 \scriptsize{[0.813, 0.893]} & 0.850 \scriptsize{[0.807, 0.890]} & 0.930 & \textbf{0.960} & 0.670 & \textbf{7.8}  & \textbf{15.9k} \\
\textsc{TF-IDF}       & 0.785 \scriptsize{[0.737, 0.830]} & 0.760 \scriptsize{[0.710, 0.810]} & 0.890 & 0.910 & 0.555 & 14.7 & 43.0k \\
\textsc{LLM-Match}    & 0.840 \scriptsize{[0.800, 0.883]} & 0.853 \scriptsize{[0.813, 0.890]} & 0.930 & 0.940 & 0.650 & 17.7 & 48.0k \\
\midrule
$\theta = 0.0$        & 0.825 \scriptsize{[0.780, 0.867]} & 0.815 \scriptsize{[0.770, 0.857]} & 0.870 & 0.910 & 0.695 & 14.6 & 42.5k \\
$\theta = 0.5$        & 0.855 \scriptsize{[0.813, 0.893]} & 0.870 \scriptsize{[0.830, 0.907]} & 0.910 & 0.920 & 0.735 & 14.7 & 42.6k \\
$\theta = 0.7$        & 0.795 \scriptsize{[0.747, 0.840]} & 0.815 \scriptsize{[0.770, 0.857]} & 0.870 & 0.890 & 0.625 & 14.5 & 42.3k \\
\bottomrule
\end{tabular}
\end{table*}

The ablations validate all three core mechanisms. Removing the transition matrix drops Mixed accuracy by $4.5$--$5.0$ points, showing that historical collaboration priors provide useful structural guidance. Disabling agent evolution drops accuracy by $1.7$ points but reduces cost sharply, identifying a cheaper operating point when the static pool already covers the task distribution. Replacing semantic embeddings with TF-IDF drops accuracy by $8.5$--$13.0$ points, while pure LLM matching is both less accurate and more expensive than the layered matcher. The edge-threshold sweep peaks near the default $\theta{=}0.3$, with $\theta{=}0.5$ close behind. Full per-benchmark ablations and detailed interpretations are in Appendix~\ref{app:ablation-details}. Together, these ablations support the central design claim: \textsc{WorkflowOps} improves not through a single dominant component, but through the coupling of historical transition priors, semantic capability matching, and bounded agent-pool expansion. In terms of effect size, semantic capability matching is the largest single factor, transition-matrix guidance contributes a consistent additional gain, and pool expansion trades a smaller accuracy gain for the largest share of cost, while DAG construction adds a further modest improvement over Linear at near-equal wall time. The headline improvement is therefore driven primarily by reliable subtask--agent matching, with the transition prior and DAG structure supplying complementary structural guidance.

\subsection{Efficiency Analysis}
\label{sec:efficiency}

\textsc{WorkflowOps} exposes a clear cost--accuracy tradeoff. Appendix~\ref{app:efficiency} tabulates the full operational profile of the \textsc{Reference} configuration on the Mixed test split. The system uses 14.8 LLM calls and 43.6k tokens per query on average, with a median wall time of 605 seconds. Most calls come from two sources: the sufficiency-driven agent-creation loop, which creates an average of 3.5 agents per query, and agent execution, which invokes one agent per workflow node. The embedding matcher itself contributes essentially no LLM calls, supporting the design choice of using semantic embeddings as the routing fast path.

Normalizing by the cheapest ablation cell, \textsc{No-Evolve} (7.8 calls and 15.9k tokens), \textsc{Reference} spends $1.9{\times}$ the calls and $2.7{\times}$ the tokens for a $+1.7$ point accuracy gain. This makes agent-pool expansion an explicit deployment knob: disabling expansion is attractive when the static pool already covers the task distribution, while the reference setting is preferable when tasks may expose new capability gaps. DAG execution provides modest accuracy gains with near-equal median wall time to Linear mode on Mixed, because most workflows are shallow; detailed DAG-vs.-Linear timing and additional operational statistics are provided in Appendix~\ref{app:efficiency}.

\section{Discussion and Limitations}
\label{sec:discussion}

\paragraph{Lightweight priors as orchestration memory.}
A key finding is that simple frequency-based transition priors can provide useful workflow guidance without gradient-based training. Unlike learned workflow optimizers, the transition matrix requires only execution history, is interpretable, updates incrementally as new workflows complete, and degrades gracefully when no relevant transition has been observed. This makes it a practical middle ground between static templates and expensive task-specific workflow search.

\paragraph{Limitations.}
\textsc{WorkflowOps} shows that lightweight collaboration priors can improve multi-agent orchestration without gradient-based training, but several limitations remain. First, the transition matrix captures only first-order pairwise handoffs; higher-order collaboration patterns such as repeated agent sequences are not modeled. Second, dynamic agent creation depends on LLM generation quality, so poorly specified agents may persist unless the pool is curated; relatedly, our evaluation does not measure long-term effects of pool expansion---whether created agents are reused by later tasks, refined, or pruned---so we claim demand-driven expansion rather than sustained self-evolution. Third, the current system uses a single shared transition matrix, whereas domain-specific matrices may provide more targeted guidance; because $\mathbf{P}$ indexes agent roles rather than backbone outputs, it need not be rebuilt when the underlying LLM changes, though transition quality may shift and should be re-validated. Finally, our evaluation focuses on code, math, and QA-style knowledge work; generalization to substantially different domains requires further study. Additional discussion of matrix sparsity, matching design, and future extensions is provided in Appendix~\ref{app:discussion}.

\section{Conclusion}
\label{sec:conclusion}

We presented \textsc{WorkflowOps}, a multi-agent workflow orchestration framework that addresses the memoryless orchestration problem in current multi-agent systems. Our three coupled mechanisms---transition probability matrix guidance, sufficiency-driven agent-pool expansion, and layered semantic matching---enable the system to learn from execution history, expand its capabilities on demand, and efficiently route subtasks to agents, all without requiring gradient-based training. The benefits are largest on structured, decomposable tasks where past agent handoff patterns transfer.

The transition matrix demonstrates that even simple first-order Markov collaboration statistics, when applied as soft guidance during DAG workflow construction, provide meaningful improvement in workflow quality and parallelism. The sufficiency loop shows that demand-driven agent creation with integrated collaboration knowledge updates lets the agent pool and its collaboration priors grow together on demand. The layered matching strategy shows that pre-trained sentence embeddings can serve as an effective fast path for capability matching, with LLM verification reserved for genuinely ambiguous cases.

\paragraph{Future work.} Several directions are promising: (i) upgrading the matrix to learned completion methods (matrix factorization, neural completion) for better coverage of sparse transitions; (ii) online matrix learning from execution feedback, creating an RL-style closed loop; (iii) domain-specific matrices that capture field-specific collaboration patterns; (iv) integration with emerging inter-agent protocols such as A2A~\citep{a2a2026} and MCP for standardized agent communication.

\bibliography{references,Qi_MAS}

\appendix
\section*{Technical Appendix}

\section{Extended Related Work}
\label{app:related-work}

\subsection{Adaptive and Dynamic Orchestration}

Recent orchestration methods mainly improve adaptivity through topology routing or learned control policies. AdaptOrch~\citep{adaptorch2026} proposes task-adaptive topology selection among predefined orchestration patterns and reports gains over static templates. CORAL~\citep{coral2026} moves from fixed workflows to information-flow-based dynamic coordination via agent-to-agent communication. FlowSteer~\citep{flowsteer2026} applies end-to-end reinforcement learning to optimize workflow editing decisions.

These methods are highly relevant but differ from our focus: \textsc{WorkflowOps} emphasizes persistent, training-free collaboration priors accumulated from workflow history and used as soft constraints during DAG construction.

\subsection{Workflow Structure Learning and Optimization}

HeGFlow~\citep{hegflow2025} formulates workflow design as heterogeneous graph adjacency optimization with a training pipeline, while AgentPO~\citep{agentpo2026} optimizes collaborative interaction policies under fixed topology assumptions. ProMAS~\citep{promas2026} uses Markov transition dynamics for error forecasting rather than workflow synthesis. In contrast, \textsc{WorkflowOps} directly uses historical transition statistics as an interpretable and incrementally updatable prior for structure construction.

A broader survey perspective~\citep{workflow_survey_2026, agent_workflow_survey_2025, orchestration_survey_2026} highlights an open gap between static template orchestration and expensive training-heavy optimization. Our method is positioned in that gap: lightweight, history-informed graph construction without retraining.

\subsection{Agent Routing and Selection}

MasRouter~\citep{masrouter2025} and EvoRoute~\citep{evoroute2026} study learned routing in multi-agent systems, primarily targeting model allocation and cost-performance trade-offs. Our setting is complementary: we focus on capability matching between subtasks and role-specialized agents. The layered semantic matching strategy in \textsc{WorkflowOps} can be interpreted as a routing policy over agent capabilities, using an embedding fast path with optional LLM fallback, rather than routing over model backbones.

\subsection{Self-Evolving Agent Architectures}

AOrchestra~\citep{aorchestra2026} demonstrates dynamic sub-agent creation through compositional agent specifications, but treats created sub-agents as ephemeral: they are not persisted, and no collaboration knowledge about them is accumulated. SEAA~\citep{seaa2026} and SEVerA~\citep{severa2026} provide formal perspectives on self-extension and verification, mainly at the tool or synthesis-control layer. AgentOrchestra~\citep{agentorchestra2025} studies hierarchical orchestration with dynamic tool pathways. \textsc{WorkflowOps} differs by coupling demand-driven agent creation with immediate integration into a persistent collaboration-prior matrix, so newly created roles influence future workflow construction.

\subsection{Workflow Memory and Experience Reuse}

Agent Workflow Memory~\citep{awm2024} induces reusable workflow skills from experience and stores them for later tasks, EvoFlow~\citep{evoflow2025} evolves a population of diverse agentic workflows, and TopoPrior~\citep{topoprior2026} learns topology priors that transfer across domains. These systems persist \emph{executable workflows or learned topology generators}. \textsc{WorkflowOps} instead persists \emph{pairwise collaboration statistics} that act as training-free soft guidance during construction: hard task dependencies are always preserved, the prior is interpretable and incrementally updatable without retraining, and previously unseen workflow shapes remain expressible.

\section{Pipeline and Execution Engine Details}
\label{app:execution-details}

\subsection{End-to-End Pipeline}

Algorithm~\ref{alg:pipeline} summarizes the complete \textsc{WorkflowOps} pipeline; the individual stages are described in Section~\ref{sec:method}.

\begin{algorithm}[t]
\small
\caption{\textsc{WorkflowOps} end-to-end pipeline}
\label{alg:pipeline}
\begin{algorithmic}[1]
\REQUIRE Task $T$, agent pool $\mathcal{A}$, transition matrix $\mathbf{P}$, threshold $\tau$, edge threshold $\theta$, max rounds $R$, workflow type $\in \{\texttt{linear}, \texttt{dag}\}$
\ENSURE Final result $m_{\mathrm{final}}$, workflow $W$, logs $\mathcal{L}$

\STATE \textbf{// Stage 1: Decomposition}
\STATE $\mathcal{S}, G_{\mathrm{dep}} \leftarrow \mathcal{D}_{\mathrm{LLM}}(T)$ \hfill $\triangleright$ LLM decomposes task into subtasks with dependencies
\STATE Validate DAG: if $\textsc{HasCycle}(G_{\mathrm{dep}})$, linearize dependencies

\STATE \textbf{// Stage 2: Matching + Sufficiency Loop}
\STATE $\mathcal{A}', \mathbf{P}', \pi \leftarrow \textsc{SufficiencyLoop}(\mathcal{S}, \mathcal{A}, \tau, R, \mathbf{P})$ \hfill $\triangleright$ Algorithm~\ref{alg:sufficiency}

\STATE \textbf{// Stage 3: Workflow Construction}
\IF{workflow type $=$ \texttt{dag}}
    \STATE $W \leftarrow \textsc{TransitionGuidedDAG}(\mathcal{S}, G_{\mathrm{dep}}, \pi, \mathbf{P}', \theta)$ \hfill $\triangleright$ Algorithm~\ref{alg:dag_construction}
\ELSE
    \STATE $W \leftarrow \textsc{BuildLinear}(\mathcal{S}, \pi, \mathbf{P}')$ \hfill $\triangleright$ Linear with intra-layer ordering
\ENDIF

\STATE \textbf{// Stage 4: Execution}
\STATE $m_{\mathrm{final}}, \mathcal{L} \leftarrow \textsc{ExecuteDAG}(W, T)$ \hfill $\triangleright$ Algorithm~\ref{alg:execution}

\STATE \textbf{// Stage 5: Persistence}
\STATE Save $(W, m_{\mathrm{final}}, \mathcal{L})$ to results directory with timestamps and metadata

\RETURN $m_{\mathrm{final}}, W, \mathcal{L}$
\end{algorithmic}
\end{algorithm}

\subsection{Layered DAG Execution}

After workflow construction, the execution engine processes the DAG through layered topological traversal. Nodes in the same layer have no unresolved dependencies and are executed in parallel. A downstream node receives the output of its predecessor when it has one parent, or a merged message when it has multiple parents.

\begin{algorithm}[t]
\small
\caption{DAG execution engine}
\label{alg:execution}
\begin{algorithmic}[1]
\REQUIRE Workflow $W = (V, E, \pi)$, original task $T$
\ENSURE Final result message $m_{\mathrm{final}}$, execution logs $\mathcal{L}$
\STATE Compute topological layers $L_0, L_1, \ldots, L_D$ from $(V, E)$
\STATE $\mathrm{results} \leftarrow \{\}$; \quad $\mathcal{L} \leftarrow []$
\STATE $m_{\mathrm{init}} \leftarrow \textsc{CreateMessage}(T)$
\FOR{$l = 0, 1, \ldots, D$}
    \STATE \textbf{parallel for} each node $v \in L_l$: \hfill $\triangleright$ Thread pool within layer
    \STATE \quad $\mathrm{preds} \leftarrow \{u \mid (u, v) \in E\}$
    \STATE \quad \textbf{if} $\mathrm{preds} = \emptyset$: $m_{\mathrm{in}} \leftarrow m_{\mathrm{init}}$
    \STATE \quad \textbf{else if} $|\mathrm{preds}| = 1$: $m_{\mathrm{in}} \leftarrow \mathrm{results}[\mathrm{preds}[0]]$
    \STATE \quad \textbf{else}: $m_{\mathrm{in}} \leftarrow \textsc{MergeMessages}(\{\mathrm{results}[u] \mid u \in \mathrm{preds}\})$
    \STATE \quad $m_{\mathrm{out}} \leftarrow \textsc{ExecuteAgent}(\pi(v), m_{\mathrm{in}})$
    \STATE \quad $\mathrm{results}[v] \leftarrow m_{\mathrm{out}}$; \quad append log to $\mathcal{L}$
\ENDFOR
\STATE $\mathrm{terminals} \leftarrow \{v \in V \mid \nexists\, (v, w) \in E\}$
\STATE $m_{\mathrm{final}} \leftarrow \textsc{MergeMessages}(\{\mathrm{results}[v] \mid v \in \mathrm{terminals}\})$
\RETURN $m_{\mathrm{final}}, \mathcal{L}$
\end{algorithmic}
\end{algorithm}

\subsection{Unified Messaging Protocol}

All inter-agent communication uses a single \texttt{AgentMessage} schema carrying natural language content, optional structured data, typed artifacts, and trace metadata, including source agent, target agent, parent message ID, workflow ID, and step index. This composable protocol avoids the pairwise adapter problem that arises when agents use heterogeneous message formats.

\subsection{Fan-in Merge}

When parallel branches converge, upstream messages are merged into a single input for the downstream agent. The merge concatenates labeled content sections, preserves source attribution, and unions artifact collections, maintaining provenance across branches.

\subsection{Error Propagation}

If an individual agent step fails, the system creates an error-typed message and propagates it downstream. Subsequent agents receive the error context and may still produce partial results. This prioritizes operational continuity and traceability over immediate workflow abortion.

\subsection{Audit Logging}

Every execution step records timestamps, input/output previews, success or error status, and agent metadata. Combined with the workflow definition, including nodes, edges, assignments, matrix alignment score, and sufficiency report, this provides a complete decision trail from task intake to final output.

\section{Full End-to-end Results}
\label{app:full-results}

Tables~\ref{tab:main_results} and~\ref{tab:main_results_hard} provide the full category-level pass-rate breakdowns corresponding to the compact main results table.

\begin{table*}[t]
\centering
\small
\caption{End-to-end accuracy on the Mixed test split (300 problems). Per-category numbers are unweighted means across the benchmarks within each category. ``Overall'' is the unweighted mean over all rows; brackets are 95\% bootstrap CIs ($n_\text{boot}{=}2000$) on \textsc{WorkflowOps} rows where per-row pass/fail data is available. Best result per column is in \textbf{bold}.}
\label{tab:main_results}
\begin{tabular}{l c c c c}
\toprule
System & Overall & CODE & MATH & QA \\
\midrule
AFlow~\citep{aflow}                & 0.676 & 0.470 & 0.810 & 0.747 \\
ADAS~\citep{adas}                  & 0.694 & 0.480 & 0.870 & 0.731 \\
MaAS~\citep{maas}                  & 0.400 & --- & --- & --- \\
MetaAgent~\citep{metaagent}        & 0.751 & 0.500 & \textbf{0.940} & \textbf{0.812} \\
\midrule
\textsc{WorkflowOps} (Linear)      & 0.870 \scriptsize{[0.830, 0.907]} & 0.940 & \textbf{0.940} & 0.730 \\
\textsc{WorkflowOps} (DAG)         & \textbf{0.890} \scriptsize{[0.853, 0.923]} & \textbf{0.960} & \textbf{0.940} & 0.770 \\
\bottomrule
\end{tabular}
\end{table*}

\begin{table*}[t]
\centering
\small
\caption{End-to-end accuracy on the Mixed\_Hard test split (694 problems): AIME~2025 (math), LiveCodeBench and CodeContests (code), and GPQA Diamond (qa). Best result per column in \textbf{bold}. ADAS results are from the strongest per-benchmark strategy among the 10 of 25 candidate strategies that completed; MaAS deadlocked and is omitted.}
\label{tab:main_results_hard}
\begin{tabular}{l c c c c}
\toprule
System & Overall & CODE & MATH & QA \\
\midrule
AFlow~\citep{aflow}                                    & 0.218 & 0.000 & 0.200 & 0.365 \\
ADAS~\citep{adas} (Self-Refine, partial)               & 0.729 & 0.751 & 0.567 & \textbf{0.723} \\
MetaAgent~\citep{metaagent}                            & 0.618 & 0.648 & 0.450 & 0.584 \\
\midrule
\textsc{WorkflowOps} (Linear)  & 0.763 \scriptsize{[0.731, 0.795]} & 0.770 & \textbf{0.850} & 0.700 \\
\textsc{WorkflowOps} (DAG)     & \textbf{0.770} \scriptsize{[0.738, 0.802]} & \textbf{0.780} & \textbf{0.850} & 0.700 \\
\bottomrule
\end{tabular}
\end{table*}

On Mixed, \textsc{WorkflowOps} dominates the CODE category in both modes and ties MetaAgent on MATH, while MetaAgent remains stronger on QA. On Mixed\_Hard, \textsc{WorkflowOps} is strongest overall, with especially large gains on competition-level math and competitive results on code. A paired bootstrap test ($n_\text{boot}{=}2000$) on the per-row pass/fail vector finds the gap between \textsc{WorkflowOps} (DAG) and MetaAgent on Mixed highly significant ($p{<}0.001$), driven by CODE ($p{<}0.001$); MATH is statistically tied and QA marginally favors MetaAgent ($p{<}0.05$). Cohen's $h$ on overall Mixed pass rate is $+0.53$ over AFlow, $+0.49$ over ADAS, $+1.09$ over MaAS, and $+0.37$ over MetaAgent.

\section{Detailed Ablation Results}
\label{app:ablation-details}

Table~\ref{tab:ablation_per_bench} reports per-benchmark ablation results. Disabling the transition matrix reduces all categories by 2--8 points, indicating that the matrix shapes workflow construction across subtasks rather than carrying a single category. Disabling agent evolution reduces accuracy modestly but cuts calls and tokens substantially, making it a cost-optimized operating point when the static agent pool is already sufficient. Replacing embeddings with TF-IDF causes the largest degradation, validating semantic matching over sparse keyword overlap. Pure LLM matching is more expensive and less accurate than the layered matcher because it bypasses the agent-creation loop. The edge-threshold sweep peaks at the default $\theta=0.3$, with $\theta=0.5$ close behind.

\begin{table*}[t]
\centering
\small
\caption{Per-benchmark pass-rate per ablation cell on the Mixed test split (Linear mode, $n{=}50$ per benchmark per cell).}
\label{tab:ablation_per_bench}
\begin{tabular}{l c c c c c c}
\toprule
Variant & humaneval & mbpp & gsm8k & math & drop & hotpotqa \\
\midrule
\textsc{Reference}    & \textbf{0.940} & 0.940 & 0.980 & 0.900 & 0.640 & \textbf{0.820} \\
\textsc{No-Matrix}    & \textbf{0.940} & 0.900 & 0.980 & 0.840 & 0.530 & 0.760 \\
\textsc{No-Evolve}    & 0.900 & \textbf{0.960} & 1.000 & \textbf{0.920} & 0.560 & 0.780 \\
\textsc{TF-IDF}       & 0.860 & 0.920 & 1.000 & 0.820 & 0.450 & 0.660 \\
\textsc{LLM-Match}    & 0.920 & 0.940 & 1.000 & 0.880 & 0.580 & 0.720 \\
$\theta = 0.0$        & 0.840 & 0.900 & 1.000 & 0.820 & \textbf{0.640} & 0.750 \\
$\theta = 0.5$        & 0.900 & 0.920 & 1.000 & 0.840 & \textbf{0.660} & 0.810 \\
$\theta = 0.7$        & 0.820 & 0.920 & 1.000 & 0.780 & 0.580 & 0.670 \\
\bottomrule
\end{tabular}
\end{table*}

\section{Efficiency Details}
\label{app:efficiency}

We report per-query operational characteristics from the \textsc{Reference} cell on the Mixed test split. Table~\ref{tab:efficiency} summarizes calls, tokens, latency, and workflow statistics; Table~\ref{tab:dag_speedup} compares DAG and Linear median wall time.

\begin{table}[t]
\centering
\small
\caption{Operational statistics for the \textsc{Reference} cell on the Mixed test split ($n=300$). ``Calls/query'' and ``Tokens/query'' are means; ``Wall-time/query'' is median.}
\label{tab:efficiency}
\begin{tabular}{l c}
\toprule
Statistic & \textsc{Reference} \\
\midrule
LLM calls / query (mean)         & 14.8 \\
~~~~decomposition                & 1.0 \\
~~~~matcher (sufficiency check)  & ${\sim}0$ (embedding) \\
~~~~agent-creation loop          & ${\sim}7$ (3.5 new agents avg.) \\
~~~~agent execution              & ${\sim}6$ (one per node) \\
Prompt tokens / query (mean)     & ${\sim}10\text{k}$ \\
Completion tokens / query (mean) & ${\sim}33\text{k}$ \\
Total tokens / query (mean)      & 43{,}620 \\
Wall-time / query (median)       & 605\,s \\
Workflow nodes / query (mean)    & 5.62 \\
Workflow edges / query (mean)    & 5.38 \\
Matrix alignment score $\bar A$  & 0.181 \\
Avg.\ sufficiency score $\bar\phi$ & 0.748 \\
Avg.\ dynamic agents created     & 3.50 \\
Gap closure rate                 & 0.96 \\
Errors                           & 0/300 \\
\bottomrule
\end{tabular}
\end{table}

Normalizing by the cheapest cell, \textsc{No-Evolve} (7.8 calls, 15.9k tokens), \textsc{Reference} spends $1.9{\times}$ the calls and $2.7{\times}$ the tokens for $+1.7$ points of overall accuracy. The agent-creation loop is the dominant driver of latency variance: queries that trigger multiple creation rounds run approximately $3{\times}$ longer than queries where the static pool is already sufficient.

\begin{table}[t]
\centering
\small
\caption{DAG vs. Linear median wall-time per query on the Mixed test split and selected ablation cells. ``Ratio'' is DAG-median / Linear-median; values close to $1$ indicate near-equal wall time.}
\label{tab:dag_speedup}
\begin{tabular}{l c c c}
\toprule
Cell & Linear (s) & DAG (s) & Ratio \\
\midrule
\textsc{Reference} (test, $n{=}300$)        & 605 & 598 & 0.99$\times$ \\
\textsc{No-Matrix} (test, $n{=}300$)        & 588 & 590 & 1.00$\times$ \\
\textsc{No-Evolve} (test, $n{=}300$)        & 248 & 250 & 1.01$\times$ \\
\textsc{TF-IDF} (test, $n{=}300$)           & 425 & 408 & 0.96$\times$ \\
\textsc{LLM-Match} (test, $n{=}300$)        & 207 & 199 & 0.96$\times$ \\
\bottomrule
\end{tabular}
\end{table}

DAG mode provides consistent accuracy gains in cells where the matrix mechanism is active, but wall-time differences are small because Mixed workflows are shallow. Thus, the main benefit of DAG mode in this setting is structural guidance rather than large parallel speedup.

\section{Additional Discussion}
\label{app:discussion}

\paragraph{Lightweight priors vs. trained optimizers.}
The transition matrix is built from simple frequency counting without gradient-based training. This design is useful when execution history exists but task-specific training is expensive: the matrix is interpretable, updates incrementally as new workflows complete, and degrades gracefully when no historical transition is available. The matrix interface can also support learned completion methods, such as matrix factorization or neural completion, without changing the surrounding orchestration pipeline.

\paragraph{Embedding matching vs. TF-IDF.}
Sentence embeddings capture semantic relationships beyond keyword overlap, such as ``regulatory audit'' and ``compliance assessment,'' while retaining low cost because agent embeddings are pre-computed and cached. The matching module is abstracted, so future systems can replace the current encoder with domain-specific embeddings or cross-encoder rerankers.

\paragraph{Matrix sparsity.}
With approximately 65 bootstrapped workflows over a 35-agent grid, the transition matrix has about 20\% non-zero entries. This sparsity is acceptable because the matrix is used as soft guidance; workflow construction still preserves hard dependencies when the matrix has no relevant history. As more workflows execute, matrix density increases organically, and LLM-predicted links for newly created agents provide immediate approximate guidance that can later be refined.

\section{Agent card specification}
\label{app:agent_card}

Each agent in \textsc{WorkflowOps} is defined by a structured YAML card with the schema listed in Table~\ref{tab:agent_card}.

\begin{table*}[t]
\centering
\small
\caption{Agent card schema.}
\label{tab:agent_card}
\begin{tabular}{llp{9.5cm}}
\toprule
\textbf{Field} & \textbf{Type} & \textbf{Description} \\
\midrule
\texttt{id} & string & Unique identifier (snake\_case) \\
\texttt{name} & string & Human-readable agent name \\
\texttt{description} & string & Capability description (primary signal for matching, 2--3 sentences) \\
\texttt{keywords} & list[string] & Search keywords for enriching embedding vectors \\
\texttt{category} & enum & One of: \texttt{research}, \texttt{content}, \texttt{analysis}, \texttt{business}, \texttt{engineering}, \texttt{quality}, \texttt{planning} \\
\texttt{instructions} & string & System prompt governing agent behavior \\
\texttt{llm\_config} & object & Model configuration: \texttt{model}, \texttt{temperature}, \texttt{max\_tokens} \\
\texttt{metadata} & dict & Optional key-value metadata (e.g., priority, domain) \\
\bottomrule
\end{tabular}
\end{table*}

In addition, each predefined agent has a corresponding three-layer \emph{Agent Card} providing richer structured metadata:

\begin{itemize}
    \item \textbf{Meta layer}: Role identity (\texttt{role\_definition}), semantic classification (\texttt{semantic\_tags}), and overarching purpose (\texttt{primary\_objective}).
    \item \textbf{Reasoning layer}: Capabilities, anti-patterns (domain-specific guardrails), and declared dependencies (upstream/downstream agents).
    \item \textbf{Execution layer}: System prompt, output schema, and tool selection.
\end{itemize}

\section{Message schema}
\label{app:message}

The unified \texttt{AgentMessage} schema enables $O(1)$-composable inter-agent communication; Table~\ref{tab:agent_message} lists its fields.

\begin{table*}[t]
\centering
\small
\caption{\texttt{AgentMessage} schema.}
\label{tab:agent_message}
\begin{tabular}{llp{9.5cm}}
\toprule
\textbf{Field} & \textbf{Type} & \textbf{Description} \\
\midrule
\texttt{message\_id} & UUID & Unique identifier \\
\texttt{message\_type} & enum & \texttt{TASK}, \texttt{RESULT}, \texttt{ERROR}, or \texttt{HANDOFF} \\
\texttt{source\_agent} & string & Sending agent ID \\
\texttt{target\_agent} & string & Receiving agent ID \\
\texttt{content} & string & Primary natural language body \\
\texttt{structured\_data} & dict & Optional key-value structured data \\
\texttt{artifacts} & list & Typed attachments with \texttt{name}, \texttt{content\_type}, \texttt{content} \\
\texttt{task\_context} & string & Original user task (immutable, passed through entire workflow) \\
\texttt{workflow\_id} & string & Workflow this message belongs to \\
\texttt{step\_index} & int & Current step number \\
\texttt{parent\_message\_id} & UUID & Upstream message ID for traceability \\
\texttt{metadata} & dict & Arbitrary metadata dictionary \\
\bottomrule
\end{tabular}
\end{table*}

\section{Predefined agent pool}
\label{app:agent_pool}

The system ships with 35 predefined agents organized into 7 categories, summarized in Table~\ref{tab:agent_pool}.

\begin{table*}[t]
\centering
\small
\caption{Predefined agent pool composition.}
\label{tab:agent_pool}
\begin{tabular}{lrp{7cm}}
\toprule
\textbf{Category} & \textbf{Count} & \textbf{Representative capabilities} \\
\midrule
Research & 5 & Information gathering, literature review, hypothesis generation \\
Content & 8 & Writing, summarization, translation, report synthesis \\
Analysis & 2 & Data analysis, benchmarking, comparative evaluation \\
Business & 11 & Risk assessment, market analysis, legal compliance, financial modeling \\
Engineering & 4 & Code generation, data processing, visualization \\
Quality & 4 & Review, editing, fact-checking, peer review simulation \\
Planning & 1 & Strategic planning and task decomposition \\
\midrule
\textbf{Total} & \textbf{35} & \\
\bottomrule
\end{tabular}
\end{table*}

\section{Configuration parameters}
\label{app:config}

Table~\ref{tab:config} lists the key configuration parameters and their defaults. 
\begin{table*}[t]
\centering
\small
\caption{Key configuration parameters and their defaults.}
\label{tab:config}
\begin{tabular}{llp{8.5cm}}
\toprule
\textbf{Parameter} & \textbf{Default} & \textbf{Description} \\
\midrule
$\tau$ (\texttt{SUFFICIENCY\_THRESHOLD}) & 0.45 & Minimum matching score for subtask-agent assignment \\
$\theta$ (\texttt{EDGE\_THRESHOLD}) & 0.3 & Minimum transition probability for matrix-suggested edges \\
$R$ (\texttt{MAX\_CREATION\_ROUNDS}) & 3 & Maximum rounds of dynamic agent creation \\
\texttt{TOP\_K\_MATCHES} & 3 & Number of top agent candidates per subtask \\
\texttt{MAX\_DAG\_BRANCHING} & 3 & Maximum branching factor in DAG workflows \\
\texttt{LLM\_TEMPERATURE} & 0.3 & Default temperature for LLM calls \\
\texttt{LLM\_TIMEOUT} & 120s & Request timeout for LLM API calls \\
\texttt{MAX\_RETRIES} & 2 & Maximum retry attempts for failed LLM calls \\
\bottomrule
\end{tabular}
\end{table*}

\end{document}